\documentclass{article}

\usepackage[preprint,nonatbib]{neurips_2025}
\usepackage{subfig}
\usepackage{wrapfig}
\usepackage{graphicx}
\usepackage{amsmath, amssymb}

\usepackage{algorithmic}
\usepackage{algorithm}

\usepackage[utf8]{inputenc} % allow utf-8 input
\usepackage[T1]{fontenc}    % use 8-bit T1 fonts
\usepackage{hyperref}       % hyperlinks
\usepackage{url}            % simple URL typesetting
\usepackage{booktabs}       % professional-quality tables
\usepackage{amsfonts}       % blackboard math symbols
\usepackage{nicefrac}       % compact symbols for 1/2, etc.
\usepackage{microtype}      % microtypography
\usepackage{xcolor}         % colors
\usepackage{enumitem}
\usepackage{multirow}

\title{Beyond Optimal Transport: Model-Aligned Coupling for Flow Matching}

\author{%
    Yexiong Lin \quad
    Yu Yao \quad
    Tongliang Liu\thanks{Corresponding author.}\\
    \small{Sydney AI Centre, The University of Sydney}\\
}

\begin{document}

\maketitle

\begin{abstract}
Flow Matching (FM) is an effective framework for training a model to learn a vector field that transports samples from a source distribution to a target distribution. To train the model, early FM methods use random couplings, which often result in crossing paths and lead the model to learn non-straight trajectories that require many integration steps to generate high-quality samples. To address this, recent methods adopt Optimal Transport (OT) to construct couplings by minimizing geometric distances, which helps reduce path crossings.
However, we observe that such geometry-based couplings do not necessarily align with the model's preferred trajectories, making it difficult to learn the vector field induced by these couplings, which prevents the model from learning straight trajectories. Motivated by this, we propose Model-Aligned Coupling (MAC), an effective method that matches training couplings based not only on geometric distance but also on alignment with the model's preferred transport directions based on its prediction error.
To avoid the time-costly match process, MAC proposes to select the top-$k$ fraction of couplings with the lowest error for training.
Extensive experiments show that MAC significantly improves generation quality and efficiency in few-step settings compared to existing methods. Project page: \href{https://yexionglin.github.io/mac/}{yexionglin.github.io/mac}
\end{abstract}

\section{Introduction}
Continuous-time generative models have obtained increasing attention for their ability to construct flexible mappings between simple source distributions and complex data distributions. Flow Matching (FM)~\cite{lipman2022flow, liu2022flow} is a representative method in this paradigm. It trains a model to learn a time-dependent vector field, which is a function that specifies the direction and magnitude of movement for each point in space and time. By integrating this vector field over time, FM transforms samples from a source distribution (e.g., Gaussian noise) into samples from the target data distribution. This formulation provides a principled framework for modeling continuous transformation paths between distributions, without requiring stochastic simulation or likelihood-based training.

A key step in FM is the construction of training data via couplings between source and target samples. 
These couplings define trajectories and provide supervision to guide the learning of the vector field that transports samples along the trajectories.
However, the vanilla FM method constructs these couplings by randomly matching samples from the two distributions. Though it is simple, random couplings often result in chaotic transport directions and frequent path crossings. This leads the model to learn non-straight trajectories, which in turn require many integration steps during inference to generate high-quality samples~\cite{liu2022flow}.

\begin{figure}[t] 
%\vspace{-10px}
\centering
\subfloat[Random Coupling]{\label{fig:one_a}
\includegraphics[width=0.43\textwidth]{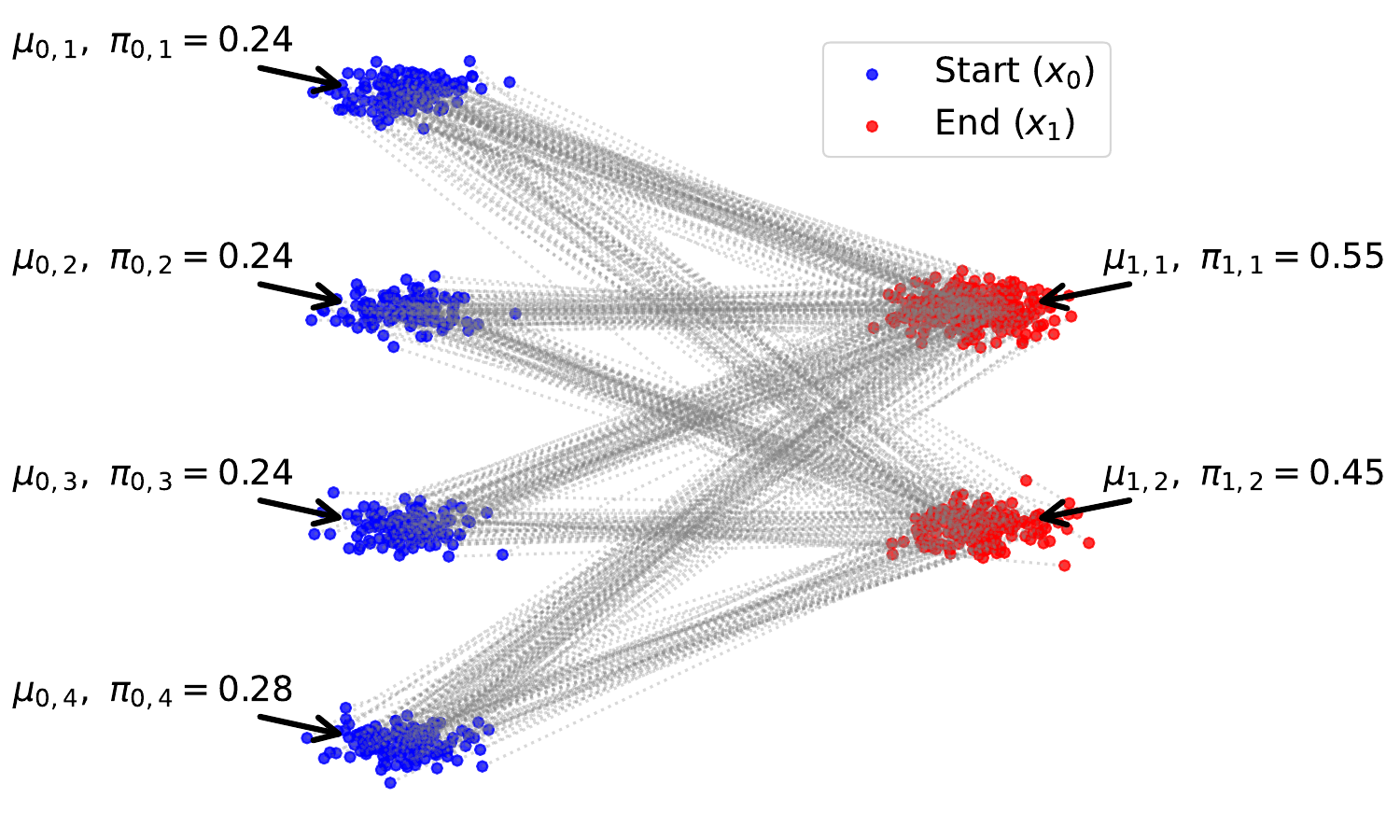}}
\!
\subfloat[Optimal Transport Coupling]{\label{fig:one_b}
\includegraphics[width=0.242\textwidth]{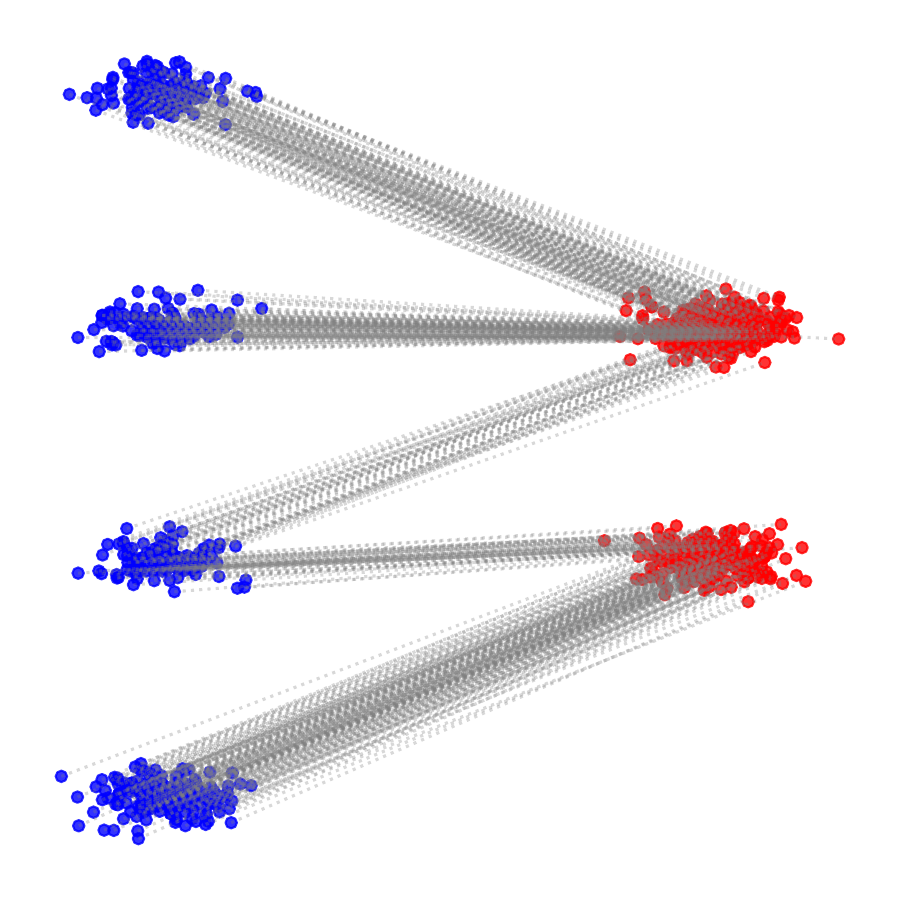}}
\quad
\subfloat[Model-Aligned Coupling (MAC)]{\label{fig:one_c}
\includegraphics[width=0.242\textwidth]{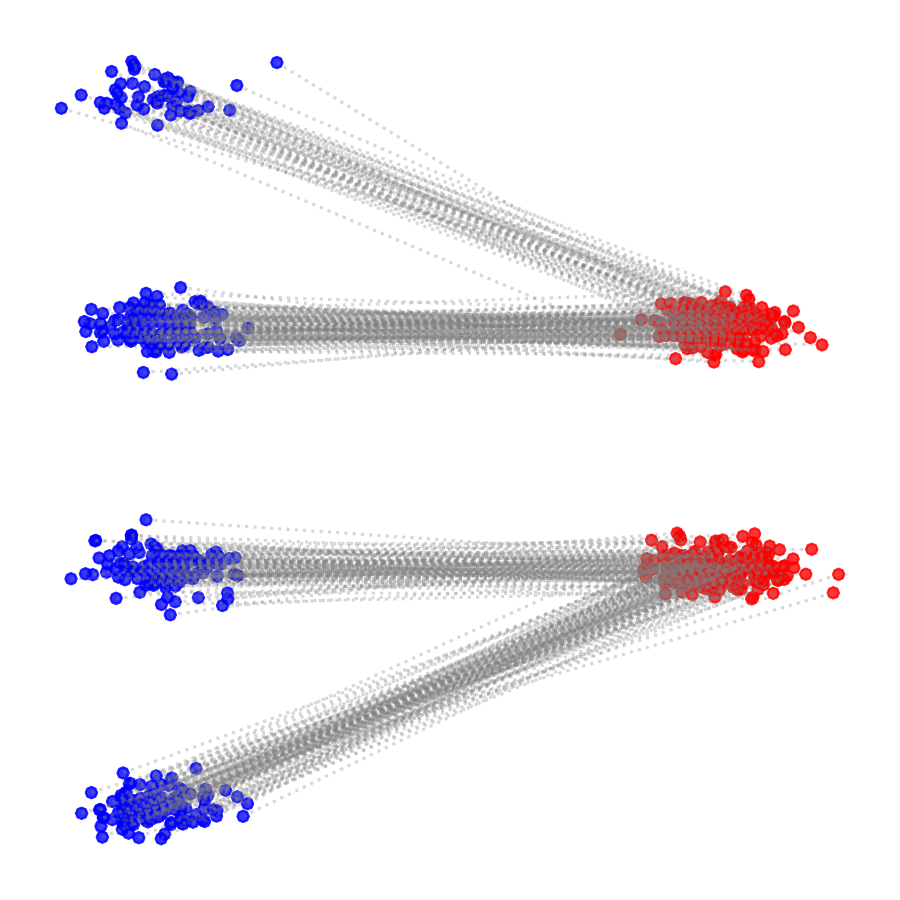}}
\\
\subfloat[Flow Matching (1 step)\label{fig:one_d}]{
\includegraphics[width=0.242\textwidth]{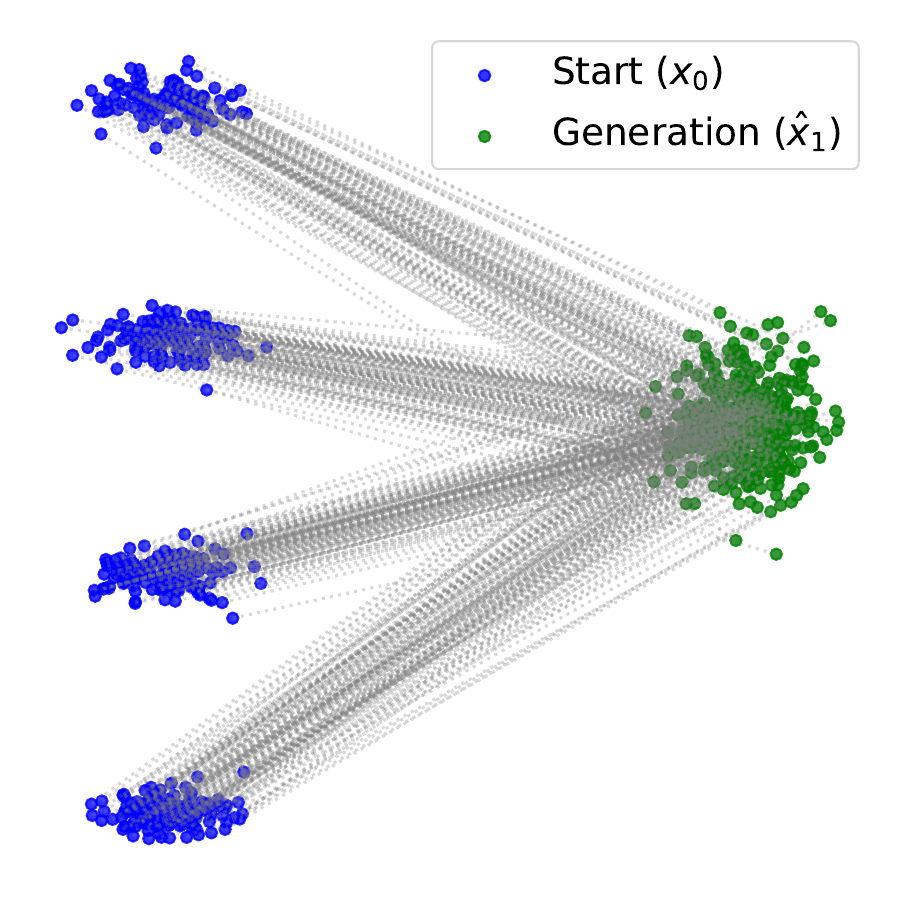}}
\!
\subfloat[OT-FM (1 step)\label{fig:one_e}]{
\includegraphics[width=0.242\textwidth]{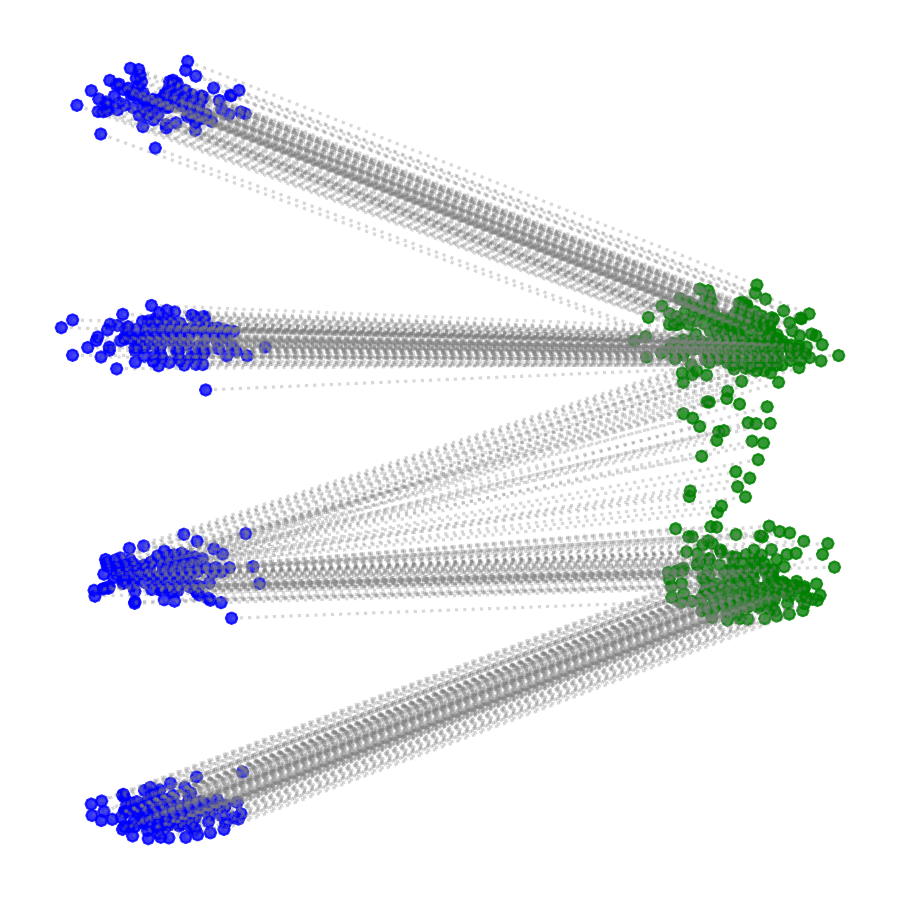}}
\!
\subfloat[Shortcut Model (1 step)\label{fig:one_f}]{
\includegraphics[width=0.242\textwidth]{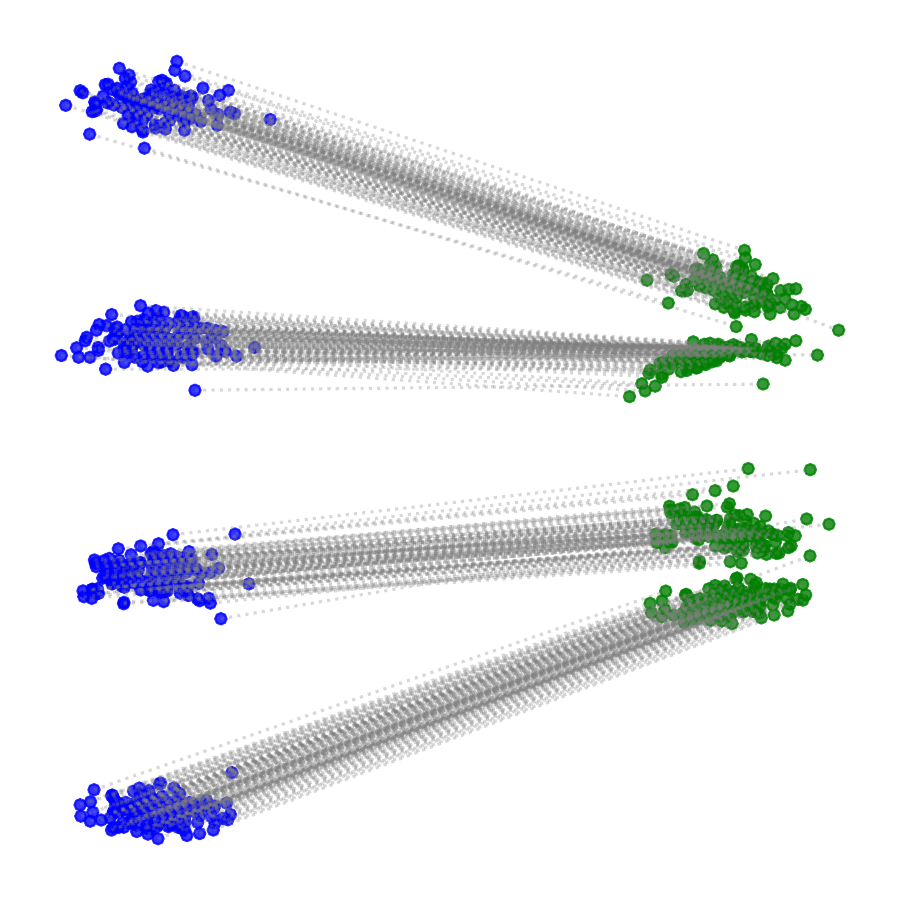}}
\!
\subfloat[MAC (1 step)\label{fig:one_g}]{
\includegraphics[width=0.242\textwidth]{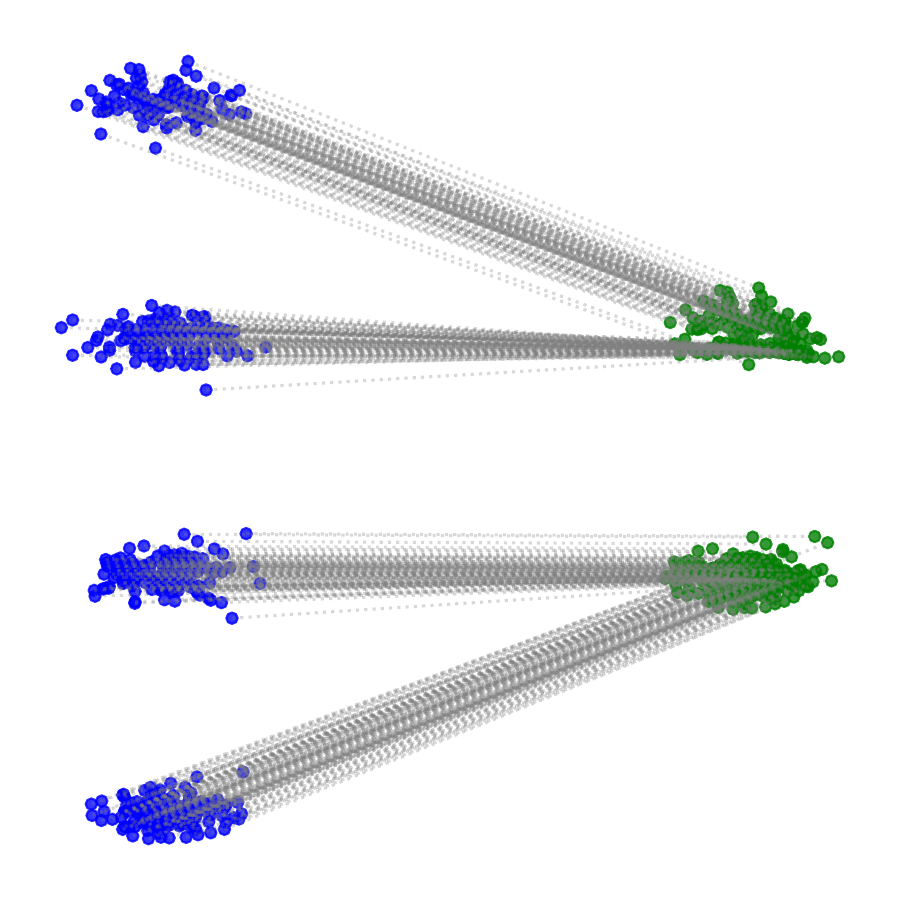}}
\caption{SubFig. (a–c) illustrate coupling strategies between a source distribution and a target distribution, defined respectively as a Gaussian mixture with four components and a Gaussian mixture with two components. Both follow the form $p(x) = \sum_{i=1}^K \pi_i \mathcal{N}(x \mid \mu_i, I)$. We compare Random Coupling, Optimal Transport (OT), and our proposed Model-Aligned Coupling (MAC). While OT improves over Random by reducing path crossings, it may still induce local ambiguity (e.g., multiple directions at $t=0$). In contrast, MAC selects couplings that better align with the model's learned vector field.
SubFig. (c-d) show one-step generated samples for different models. MAC yields significantly better sample quality with fewer integration steps compared to standard FM, OT-FM, and Shortcut models.
}
\label{fig:one}
%\vspace{-10px}
\end{figure}

To improve the quality of training supervision, recent works adopt Optimal Transport (OT) to construct couplings that minimize geometric distances (e.g., $L_2$ distance) between samples~\cite{tong2023improving,pooladian2023multisample}. By explicitly reducing the total transport cost, OT-based couplings mitigate path crossings and offer more structured transport directions compared to random coupling. However, we found that relying solely on geometric distance to match couplings is not sufficient. This strategy ignores the model's capacity, resulting in matched couplings that do not align with the learned vector field. As a result, the model cannot learn effectively, further degrading the quality of one-step generation. Fig.~\ref{fig:one}(a–b) illustrates this issue on a synthetic task that transports samples from a Gaussian mixture with four components to a Gaussian mixture with two components. While random couplings (Fig.~\ref{fig:one_a}) produce many cross paths, OT-based couplings (Fig.~\ref{fig:one_b}) yield more organized directions. Nonetheless, OT can still introduce ambiguity. For example, in Fig.~\ref{fig:one_b}, at time $t=0$, the same source Gaussian is coupled to multiple target Gaussians. This results in inconsistent transport directions originating from the same spatial location. Since the model predicts similar velocity vectors at nearby points in space and time, it cannot simultaneously capture sharply differing directions. As a result, the model regresses toward an average of the conflicting directions after training, leading to non-straight and curved trajectories. These trajectories deviate from the ideal linear transport paths, making the generative process less efficient. As a consequence, the model requires more integration steps during inference to reach the target distribution. This inefficiency is reflected in the degraded quality of one-step generation, as shown in Fig.~\ref{fig:one_e}.

Motivated by the above insight, we propose Model-Aligned Coupling (MAC), a strategy that considers not only geometric distance but also the model's capacity to learn the induced transport directions. To realize the idea, for each mini-batch, we can evaluate all possible source-target coupling and select the matching that minimizes the prediction error, as defined by the model's current flow field. Since the model's vector field evolves over time and prediction errors shift throughout training, MAC dynamically updates the selected couplings at each iteration to reflect the model's current vector field.
By aligning couplings with the model's evolving representation, MAC promotes straighter trajectories and ultimately improves generative efficiency, particularly in one-step generation.

As illustrated in Fig.~\ref{fig:one_c}, MAC selects couplings that form more coherent and learnable transport paths. When these couplings are used to train a Flow Matching model, they lead to significantly improved one-step generation quality. Fig.~\ref{fig:one}(d–g) compare one-step samples generated using different coupling strategies. Specifically, Fig.~\ref{fig:one_d} shows results from training FM with random couplings, while Fig.~\ref{fig:one_e} uses OT-based couplings. Fig.~\ref{fig:one_f} presents outputs from the Shortcut model~\cite{frans2024one}, a state-of-the-art method for one-step generation. In contrast, Fig.~\ref{fig:one_g} shows samples produced by FM trained with our proposed Model-Aligned Coupling (MAC). Across all cases, MAC yields higher-quality samples, producing more accurate and consistent generations in a single step. Notably, it outperforms both the OT-based method and the Shortcut model, demonstrating its effectiveness in bridging the gap between coupling supervision and model learnability.

The key contributions of this paper are summarized as follows:
\begin{itemize}[leftmargin=*]
    \item We introduce Model-Aligned Coupling (MAC), a simple yet effective strategy for constructing supervision couplings in Flow Matching. Unlike prior methods that rely solely on geometric distance (e.g., Optimal Transport), MAC selects couplings based on their learnability, measured by the model's prediction error, to better align with the model's current vector field.

    \item To ensure computational efficiency, MAC avoids the costly computation of all pairwise errors and global assignment, typically requiring $O(n^2 \log n)$ to $O(n^3)$ time depending on the matching algorithm (e.g., the Hungarian method), by adopting a lightweight sample-selection procedure. At each training step, it randomly samples candidate source-target couplings, computes their regression losses, and retains the top-$k$ fraction with the lowest error. This reduces the overall time complexity to $O(n \log n)$ while dynamically adapting to the model's evolving capacity.

    \item Extensive experiments on both synthetic and real-world datasets demonstrate that MAC significantly improves one-step generation quality and reduces the number of integration steps needed for high-quality sample synthesis, outperforming both Optimal Transport-based methods and state-of-the-art shortcut models.
\end{itemize}

\section{Related Work}

\paragraph{Diffusion-based Generative Models.}
Diffusion models~\cite{sohl2015deep, ho2020denoising, song2020score} have emerged as a powerful class of generative models that transform a simple prior distribution into the data distribution via a sequence of noise removal steps. These models are typically trained using either denoising score matching~\cite{song2020score} or variational inference~\cite{ho2020denoising}. More recent approaches have adopted continuous-time formulations, modeling the generative process as a stochastic differential equation (SDE) or its deterministic counterpart (ODE)~\cite{song2020score, karras2022elucidating}.

Flow Matching (FM)~\cite{lipman2022flow} offers a simulation-free alternative to traditional diffusion models. Instead of relying on iterative stochastic simulations, FM directly regresses a time-dependent velocity field that deterministically transports samples from a source distribution (e.g., Gaussian) to the target data distribution. This eliminates the need for solving differential equations during training, and allows for greater flexibility in designing couplings and supervision mechanisms.

A growing body of work has extended Flow Matching in various directions. Equivariant Flow Matching (EquiFM)~\cite{song2023equivariant} introduces symmetry-aware training objectives and hybrid probability paths for geometric data. Simulation-free Schr\"{o}dinger bridges~\cite{tong2023simulation} unify score matching and flow matching objectives to model stochastic dynamics without requiring trajectory simulations. Hierarchical Rectified Flow (HRF)~\cite{zhang2025towards} proposes a hierarchical ODE formulation to capture complex, multi-scale dynamics across domains such as position, velocity, and acceleration. Further improvements to Rectified Flows include the use of U-shaped timestep distributions and perceptual metrics such as the LPIPS-Huber loss, which enhance sample fidelity and reduce the number of function evaluations~\cite{lee2024improving}.
FM-based methods have also been adopted in applied domains. For example, simulation-based inference tasks benefit from FM's scalability and efficiency~\cite{wildberger2023flow}, while recent work explores FM for robot action prediction~\cite{song2025humeintroducingsystem2thinking, qu2025spatialvlaexploringspatialrepresentations}, demonstrating its versatility beyond image generation.

\paragraph{Coupling Strategies for Generative Models.} A key step in training Flow Matching models is the construction of sample couplings $(x_0, x_1)$. The original Flow Matching framework~\cite{lipman2022flow} relies on randomly coupling samples from the source and target distributions. However, such a strategy often leads to curved trajectories that hinder sample quality. 

To improve coupling quality, recent works have adopted Optimal Transport (OT) to construct more geometrically meaningful source-target couplings by minimizing transport costs~\cite{tong2023improving, pooladian2023multisample}. Another line of work leverages trained generative models to construct improved couplings. ReFlow~\cite{liu2022flow} proposes an iterative refinement scheme that uses previously trained models to generate new training samples, thereby progressively straightening transport paths. Rectified Diffusion~\cite{wang2024rectified} extends this idea by employing pre-trained diffusion models to synthesize samples used for coupling. The basic idea behind these methods is to replace randomly sampled couplings with model-generated ones that better reflect the underlying data geometry and simplify the learning of vector fields.

\paragraph{Efficient Sampling Generative Models.}
Sampling efficiency is a key challenge in generative modeling, particularly for diffusion-based and flow-based methods, which often require many iterative steps to produce high-quality samples. Several techniques have been proposed to reduce the number of sampling steps required in diffusion and flow-based models. Deterministic denoising diffusion implicit models (DDIM) \cite{song2020denoising} and consistency models \cite{song2023consistency} accelerate generation by modifying the inference dynamics or enforcing consistency across time steps.

Other methods, such as progressive distillation \cite{salimans2022progressive}, reduce sampling time by distilling a multi-step model into a faster one through additional training phases. Relational Diffusion Distillation (RDD) \cite{feng2024relational} proposes a novel distillation method tailored for diffusion models. By introducing cross-sample relational knowledge distillation, RDD alleviates the memory constraints induced by multiple sample interactions. DPM-Solver~\cite{lu2022dpm} introduces a fast ODE solver for diffusion probabilistic models by formulating the sampling process as solving a diffusion ODE, enabling high-quality sample generation with significantly fewer function evaluations.

Recently, shortcut models \cite{frans2024one} take a different approach by conditioning the model on both the current state and the desired step size, allowing it to directly predict the velocity required for taking large steps in the generation process This enables high-quality generation in a single or few steps through a self-consistency training objective.

\section{Model-Aligned Coupling for Flow Matching}

In this section, we present our proposed method, Model-Aligned Coupling (MAC), a general strategy for improving supervision in Flow Matching by aligning training couplings with the model's capacity to learn them. We begin by reviewing the Flow Matching framework, and then introduce MAC through its coupling construction and efficient training implementation.

\subsection{Preliminaries}

Flow Matching (FM) is a continuous-time generative modeling framework that learns a time-dependent vector field \( v_\theta(x, t) \) to transport samples from a source distribution \( p_0(x) \) (e.g., standard Gaussian) to a target data distribution \( p_1(x) \). The objective is to match a reference velocity field \( u_t(x) \) along a time-dependent probability path \( p_t(x) \) connecting \( p_0 \) and \( p_1 \) \cite{lipman2022flow}:
\begin{equation}
\label{eq:fm}
\mathcal{L}_{\mathrm{FM}}(\theta) = \mathbb{E}_{t \sim \mathcal{U}[0,1],\, x \sim p_t} \left[ \left\| v_\theta(x, t) - u_t(x) \right\|^2 \right].
\end{equation}

\paragraph{Conditional Flow Matching.} 
Computing $u_t(x)$ typically requires access to the path marginal $p_t(x)$ and, in many cases, its score function $\nabla_x \log p_t(x)$, both of which are generally intractable. To overcome this, Conditional Flow Matching (CFM) \cite{lipman2022flow} introduces a conditional path \( p_t(x \mid x_1) \) given a fixed target \( x_1 \sim p_1 \), and supervises the model using a conditional velocity field \( u_t(x \mid x_1) \). The resulting objective is:
\begin{equation}
\label{eq:cfm}
\mathcal{L}_{\mathrm{CFM}}(\theta) = \mathbb{E}_{t \sim \mathcal{U}[0,1],\, x_1 \sim p_1,\, x \sim p_t(x \mid x_1)} \left[ \left\| v_\theta(x, t) - u_t(x \mid x_1) \right\|^2 \right].
\end{equation}

Importantly, Lipman \textit{et al.} prove that the FM (Eq.~\ref{eq:fm}) and CFM (Eq.~\ref{eq:cfm}) objectives yield identical gradients with respect to \( \theta \) \cite{lipman2022flow}. This theoretical equivalence ensures that CFM provides an unbiased and tractable approach for training Flow Matching models.

\paragraph{A Practical Instantiation via Linear Interpolation.}
A practical and widely used instantiation of CFM defines the conditional path using linear interpolation between source and target samples \( x_0 \sim p_0 \) and \( x_1 \sim p_1 \), drawn jointly from a coupling \( \rho(x_0, x_1) \) \cite{liu2022flow, lee2024improving}. Each point on the interpolation path is given by:
\[
x_t = (1 - t)x_0 + t x_1,
\]
with the associated conditional velocity:
\[
u_t(x_t \mid x_1) = \frac{d x_t}{dt} = x_1 - x_0.
\]

Substituting this into the CFM objective yields a tractable training loss:
\begin{equation}
\label{eq:lin}
\mathcal{L}_{\text{lin}}(\theta) = \mathbb{E}_{t \sim \mathcal{U}[0,1],\, (x_0, x_1) \sim \rho} \left[ \left\| v_\theta(x_t, t) - (x_1 - x_0) \right\|^2 \right],
\end{equation}
which encourages the model to learn a vector field that aligns with the displacement between source and target couplings across linearly interpolated trajectories.

\subsection{Model-Aligned Coupling}

\paragraph{Selecting Learnable Couplings.} In Flow Matching, training supervision is provided through a coupling \( \rho(x_0, x_1) \) between samples from the source and target distributions. Existing works \cite{tong2023improving,pooladian2023multisample} commonly construct this coupling by minimizing geometric distances, such as the squared Euclidean norm \( \|x_0 - x_1\|^2 \). However, geometric proximity alone does not guarantee that the coupling is easy for the model to learn, which can lead to non-straight trajectories and degraded sample quality in few-step generations.

Instead of selecting couplings solely based on geometric distance, our goal is to select couplings that are better aligned with the model's current ability to fit the data. Specifically, we aim to prioritize couplings \( (x_0, x_1) \) that have lower prediction error under the current vector field \( v_\theta \). To measure whether the model can fit the data well, we employ the pairwise prediction error, which is defined as:
\begin{equation}
\label{eq:pair}
\mathcal{L}_{\mathrm{pair}}(x_0, x_1) := \mathbb{E}_{t \sim \mathcal{U}[0,1]} \left[ \mathcal{L}_{\mathrm{MSE}}(v_\theta(x_t, t), x_0, x_1) \right],
\end{equation}
where \( x_t = (1 - t)x_0 + t x_1 \), and \( \mathcal{L}_{\mathrm{MSE}}(v_\theta(x_t, t), x_0, x_1) := \left\| v_\theta(x_t, t) - (x_1 - x_0) \right\|^2 \).

Our goal is to find a coupling $\tilde{\rho}(x_0, x_1) \in \mathcal{C}(p_0, p_1)$ that minimizes the expected pairwise prediction error over the sample space:
\begin{equation}
\label{eq:rho}
\tilde{\rho} = \arg\min_{\rho \in \mathcal{C}(p_0, p_1)} \, \mathbb{E}_{(x_0, x_1) \sim \rho} \left[ \mathcal{L}_{\mathrm{pair}}(x_0, x_1) \right],
\end{equation}
where $\mathcal{C}(p_0, p_1)$ denotes the set of admissible couplings with fixed marginals $p_0$ and $p_1$. This formulation seeks couplings that are best aligned with the model's ability to represent the underlying transport field from $p_0$ to $p_1$.

\paragraph{Efficient Approximation via Top-$k$ Selection.}
While the objective in Eq.~\ref{eq:rho} defines an ideal coupling based on learnability, computing the globally optimal assignment \( \tilde{\rho} \) exactly is computationally infeasible, as it involves evaluating all possible source-target couplings and solving a global assignment problem over the joint space. 
To approximate this efficiently, we adopt a sampling-based strategy. At each training step, we draw two mini-batches $\{x_0^i\}_{i=1}^B \sim p_0$ and $\{x_1^i\}_{i=1}^B \sim p_1$, and randomly form $B$ candidate source-target couplings $\{(x_0^i, x_1^i)\}_{i=1}^B$. For each candidate coupling, we estimate its learnability using the pairwise prediction error defined in Eq.~\ref{eq:pair}.

However, calculating the full expectation in Eq.~\ref{eq:pair} over continuous time is computationally intractable. To approximate it, we evaluate the prediction error at the two endpoints of the interpolation path, $t=0$ and $t=1$, since the endpoints provide supervision signals that are most relevant to one-step generation. Specifically, for each candidate pair, we compute:
\begin{equation}
\label{eq:pair_approx}
\mathcal{L}_{\mathrm{pair}}(x_0, x_1) \approx \tfrac{1}{2} \left( \mathcal{L}_{\mathrm{MSE}}(v_\theta(x_0, 0), x_0, x_1) + \mathcal{L}_{\mathrm{MSE}}(v_\theta(x_1, 1), x_0, x_1) \right),
\end{equation}
which avoids evaluating the full interpolation trajectory while retaining useful alignment information.

We then select the top-$k$ fraction of candidate couplings with the lowest approximation error to form the learnable subset $\mathcal{S}_\theta \subset \{(x_0^i, x_1^i)\}_{i=1}^B$. Focusing on these learnable couplings, the model can learn more consistent transport directions and finally improve the generation quality in few-step generations. The subset is updated at each iteration, adapting to the evolving vector field throughout training.

\paragraph{Regularizing Flow Matching with Selected Couplings.}
Once the selected coupling set $\mathcal{S}_\theta$ is constructed, the next step is to integrate it into the training objective to guide learning more effectively. To this end, we introduce an auxiliary regularization term that prioritizes low-error, learnable transport directions identified by MAC.

Specifically, MAC can be applied as a general-purpose regularization strategy that enhances any existing Flow Matching objective. Given the selected set $\mathcal{S}_\theta$, we define the total training loss as:
\begin{equation}
\label{eq:total}
\mathcal{L}_{\mathrm{total}} = 
\mathbb{E}_{(x_0, x_1) \sim \rho} \mathcal{L}(\theta, x_0, x_1) 
+ 
\lambda \cdot \mathbb{E}_{(x_0, x_1) \sim \mathcal{S}_\theta} \mathcal{L}(\theta, x_0, x_1),
\end{equation}
where $\rho$ denotes the base coupling distribution (e.g., random), and $\lambda$ is a weighting coefficient that controls the strength of the MAC regularization. The loss function $\mathcal{L}$ can be instantiated with any flow matching variant, such as standard Flow Matching or Shortcut Models.

\begin{algorithm}[t]
\caption{Training with Model-Aligned Coupling (MAC)}
\label{alg:mac}
\begin{algorithmic}[1]
\REQUIRE Model parameters \( \theta \), batch size \( B \), selection ratio \( k \), weighting factor \( \lambda \)
\STATE \textbf{Warm-up:} Train for one epoch using randomly sampled couplings to initialize \( v_\theta \)
\FOR{each training iteration}
    \STATE Sample a batch \( \mathcal{B} = \{(x_0^i, x_1^i)\}_{i=1}^{B} \sim \rho \)
    \FOR{each pair \( (x_0, x_1) \in \mathcal{B} \)}
        \STATE Compute interpolated points \( x_t = (1 - t)x_0 + t x_1 \) for \( t = 0 \) and \( t = 1 \)
        \STATE Compute prediction error:
        \[
        \mathcal{L}_{\mathrm{pair}}(x_0, x_1) \leftarrow \tfrac{1}{2} \left(
            \| v_\theta(x_0, 0) - (x_1 - x_0) \|^2 +
            \| v_\theta(x_1, 1) - (x_1 - x_0) \|^2
        \right)
        \]
    \ENDFOR
    \STATE Identify top-\(k\) couplings with lowest errors to form \( \mathcal{S}_\theta \)
    \STATE Compute total batch loss using Eq.~\ref{eq:em}
    \STATE Update model parameters: \( \theta \leftarrow \theta - \eta \nabla_\theta \mathcal{L}_{\mathrm{empirical}} \)
\ENDFOR
\end{algorithmic}
\end{algorithm}

\paragraph{Implementation details.}
To reduce computational overhead and avoid redundant forward-backward passes when computing the regularization term in Eq.~\ref{eq:total}, we adopt a soft weighting strategy. Rather than forming a separate supervision set, we assign elevated loss weights to MAC-selected couplings. The training loss becomes a weighted average over the entire batch:
\begin{equation}
\label{eq:em}
\mathcal{L}_{\mathrm{empirical}} = \frac{1}{|\mathcal{B}|} \sum_{(x_0, x_1) \in \mathcal{B}} w(x_0, x_1) \cdot \mathcal{L}(\theta, x_0, x_1),
\end{equation}
where the weight function \( w(x_0, x_1) \) is defined as:
\[
w(x_0, x_1) = 
\begin{cases}
1 + \lambda & \text{if } (x_0, x_1) \in \mathcal{S}_\theta, \\
1 & \text{otherwise}.
\end{cases}
\]
This allows the model to receive stronger gradients from more learnable couplings, without the need to recompute separate forward passes for the selected subset.

To improve stability during early training, we introduce a one-epoch warm-up phase using randomly sampled couplings. This provides a reasonable initialization of the vector field \( v_\theta \), making prediction errors more reliable for selecting learnable couplings.

We summarize the complete training procedure in Algorithm~\ref{alg:mac}, which outlines the MAC-based coupling selection and weighting process used during training.

\section{Experiments}

In this section, we evaluate the effectiveness of Model-Aligned Coupling (MAC) through comprehensive experiments. We begin by detailing our experimental setup, including datasets, architectures, baselines, and training protocols. We then present quantitative comparisons of generation quality across varying numbers of inference steps.
We further perform ablation studies to analyze the sensitivity of MAC to key hyperparameters. Additional experimental results are provided in Appendix~\ref{app:extra} due to space constraints.

\subsection{Experimental Setup}

\paragraph{Datasets.} 
We evaluate our method on three widely used benchmarks: MNIST \cite{lecun1998gradient}, CIFAR-10 \cite{krizhevsky2009learning}, and CelebA-HQ-256 \cite{karras2017progressive}. 
\textbf{MNIST} consists of 60,000 training and 10,000 test grayscale images of handwritten digits, each of size $28\times28$ and 10 classes. We pad each image with zeros to size $32\times32$ and perform conditional generation.
\textbf{CIFAR-10} contains 50,000 training and 10,000 test images of natural objects across 10 classes, each resized to $32\times32$ RGB. We perform conditional generation.
\textbf{CelebA-HQ-256} is a high-resolution face dataset with 30,000 images. We follow common practice and perform unconditional generation in the latent space of a pretrained autoencoder.\footnote{\url{https://huggingface.co/stabilityai/sd-vae-ft-mse}}
Latent representations have shape $(4,32,32)$ and are decoded using a fixed decoder.

\paragraph{Backbone Architecture.} 
All models adopt the DiT \cite{peebles2023scalable} architecture as the backbone, with consistent configurations across baselines to ensure fair comparison. The model width, depth, and attention dimensions are tailored to each dataset and provided in detail in Appendix~\ref{app:details}.

\paragraph{Baselines.}
We compare MAC with both one-phase and two-phase training baselines to evaluate its effectiveness. Two-phase training baselines include: (1) \textbf{Progressive Distillation (PD)}~\cite{salimans2022progressive} reduces sampling steps by progressively training a student model to match the outputs of a teacher model with twice as many denoising steps;
(2) \textbf{Consistent Distillation (CD)}~\cite{song2023consistency} trains a student to produce consistent clean outputs from pairs of noisy inputs generated by a teacher model at different noise levels;
(3) \textbf{Reflow}~\cite{liu2022flow} enhances flow matching by retraining on noise-sample pairs generated from a pretrained flow model, encouraging straighter generative paths and faster sampling. We follow the standard two-phase setup and generate 50,000 synthetic training samples using a pretrained Flow Matching model to train Reflow. One-phase training baselines include:
(4) \textbf{Flow Matching (FM)}~\cite{lipman2022flow}, which learns a time-dependent vector field using randomly sampled couplings;
(5) \textbf{BatchOT}~\cite{pooladian2023multisample}, which leverages optimal transport to select low-cost couplings for vector field supervision;
(6) \textbf{Shortcut Models}~\cite{frans2024one}, which extend FM to variable-length generation by predicting velocity over arbitrary step sizes.

\paragraph{MAC with Shortcut Models.}
We now describe how MAC is integrated into the Shortcut Models framework~\cite{frans2024one}, which parameterizes the velocity field over variable-length trajectories. The model learns a velocity field \( s_\theta(x, t, d) \), where \( x \) is the position, \( t \in [0, 1] \) is the interpolation time, and \( d \in [0, 1] \) denotes the target step size. Specifically, \( d \) represents the inverse of the desired number of generative steps: \( d = 1 \) corresponds to single-step generation, while smaller values (e.g., \( d = 1/128 \)) correspond to multi-step trajectories. In the original Shortcut Models paper~\cite{frans2024one}, the model is trained with discrete step sizes \( d \in \{1/128, 1/64, \dots, 1/2, 1\} \), covering a wide spectrum from full trajectories to incremental steps. Notably, setting \( d = 0 \) recovers the standard flow matching objective.

The training objective combines flow matching at \( d = 0 \) with a self-consistency loss over larger step sizes:
\begin{equation}
\mathcal{L}_{\mathrm{SC}}(\theta, x_0, x_1) = 
\mathbb{E}_{(t,d) \sim p(t,d)} 
\left[
\underbrace{\left\| s_\theta(x_t, t, 0) - (x_1 - x_0) \right\|^2}_{\text{Flow Matching}} 
+ 
\underbrace{\left\| s_\theta(x_t, t, 2d) - s_{\text{target}} \right\|^2}_{\text{Self Consistency}}
\right],
\end{equation}
where
\[
x_t = (1 - t) x_0 + t x_1, \quad
x'_{t+d} = x_t + d \cdot s_\theta(x_t, t, d), \quad
s_{\text{target}} = \frac{1}{2} s_\theta(x_t, t, d) + \frac{1}{2} s_\theta(x'_{t+d}, t + d, d).
\]

To incorporate MAC into the Shortcut Models framework, we adapt the coupling selection process to align with the one-step generation objective. Specifically, in the pairwise prediction error defined in Eq.~\ref{eq:pair}, we replace the standard vector field $v_\theta(x_t, t)$ with the Shortcut velocity field $s_\theta(x_t, t, d=1)$, which is explicitly trained to model single-step transitions. As a result, the selected couplings in \( \mathcal{S}_\theta \) are those for which one-step transport is most learnable under the current model. Furthermore, we directly supervise the model on a subset (ratio $r$) of the couplings in \( \mathcal{S}_\theta \) using the empirical flow-matching loss with step size fixed at $d = 1$. This encourages the model to learn one-step transport trajectories that are aligned with real data, thereby enhancing generation quality under low-step settings.

\paragraph{Training Details.} 
We follow the batching strategy used in the official implementation of Shortcut Models,\footnote{\url{https://github.com/kvfrans/shortcut-models}} where each mini-batch is composed of a mixture of empirical supervision and self-consistency supervision. Specifically, only a fraction \( m = \frac{1}{8} \) of couplings in the batch are randomly assigned the self-consistency objective, while the remaining \( 1 - m \) proportion is used solely for empirical supervision. We also follow the setting in Shortcut Models to use an exponential moving average (EMA) of model parameters for both generating bootstrap targets and computing prediction error in MAC.

All models are trained with a learning rate of $5\times10^{-4}$ using default Adam optimizer parameters. The number of training epochs is 200 for MNIST and CIFAR-10, and 900 for CelebA-HQ-256. All input images are normalized to the range $[-1, 1]$ before training. We faithfully re-implement all baselines using PyTorch 2.4.1 under the same training settings to ensure fair comparison. All experiments are conducted on two NVIDIA Tesla V100 GPUs.
We use batch size 256 for MNIST and CIFAR-10, and 64 for CelebA-HQ-256. All models are trained with Adam optimizer, fixed random seeds, and exponential moving average (EMA) for evaluation. We report Fr\'echet Inception Distance (FID) computed on 10,000 samples. We compare all methods under 128, 4, and 1 diffusion step settings to evaluate sample quality and efficiency.

\subsection{Quantitative Results}

\begin{table}[t]
\centering
\caption{One-step, four-step, and full-step (128 steps) FID scores on MNIST, CIFAR-10, and CelebA-HQ-256. Lower is better. The best results within each category are highlighted in bold. MAC (ours) outperforms all other one-phase methods.}
\label{tab:fid_results}
\resizebox{\linewidth}{!}{
\begin{tabular}{l|ccc|ccc|ccc}
\toprule
\multirow{2}{*}{Method} & \multicolumn{3}{c|}{MNIST (Cond.)} & \multicolumn{3}{c|}{CIFAR-10 (Cond.)} & \multicolumn{3}{c}{CelebA-HQ (Uncond.)} \\
& 1-Step & 4-Step & 128-Step & 1-Step & 4-Step & 128-Step & 1-Step & 4-Step & 128-Step \\
\midrule
Two phase \\
\midrule
PD & 176.30 & 29.01 & \textbf{6.58} & 317.55 & 34.39 & \textbf{10.39} & 280.35 & 106.79 & 127.99 \\
CD & 68.37 & 125.55 & 181.49 & 120.72 & 94.45 & 119.65 & 143.47 & \textbf{23.61} & 39.26 \\
Reflow & \textbf{61.23} & \textbf{16.22} & 11.78 & \textbf{21.75} & \textbf{16.18} & 14.97 & \textbf{35.85} & 23.62 & \textbf{19.79} \\
\midrule
One phase \\
\midrule
Flow Matching & 169.97 & 28.99 & 7.13 & 324.04 & 36.85 & 11.05 & 284.99 & 65.05 & 15.25 \\
BatchOT       & 167.67 & 30.69 & 6.82 & 314.93 & 36.64 & 11.27 & 282.14 & 64.95 & 15.12 \\
Shortcut      &  75.03 & 25.80 & 7.22 &  36.19 & \textbf{17.14} & 11.96 & 32.20 & 17.76 & \textbf{11.16} \\
MAC (ours)    &  \textbf{68.21} & \textbf{22.47} & \textbf{6.69} &  \textbf{35.47} & 19.14 & \textbf{10.44} & \textbf{26.48} & \textbf{13.84} & 11.40 \\
\bottomrule
\end{tabular}
}
\end{table}

\begin{figure}[t]
\vspace{-10px}
\centering
\subfloat[Hyper-parameter $k$]{
\includegraphics[width=0.32\textwidth]{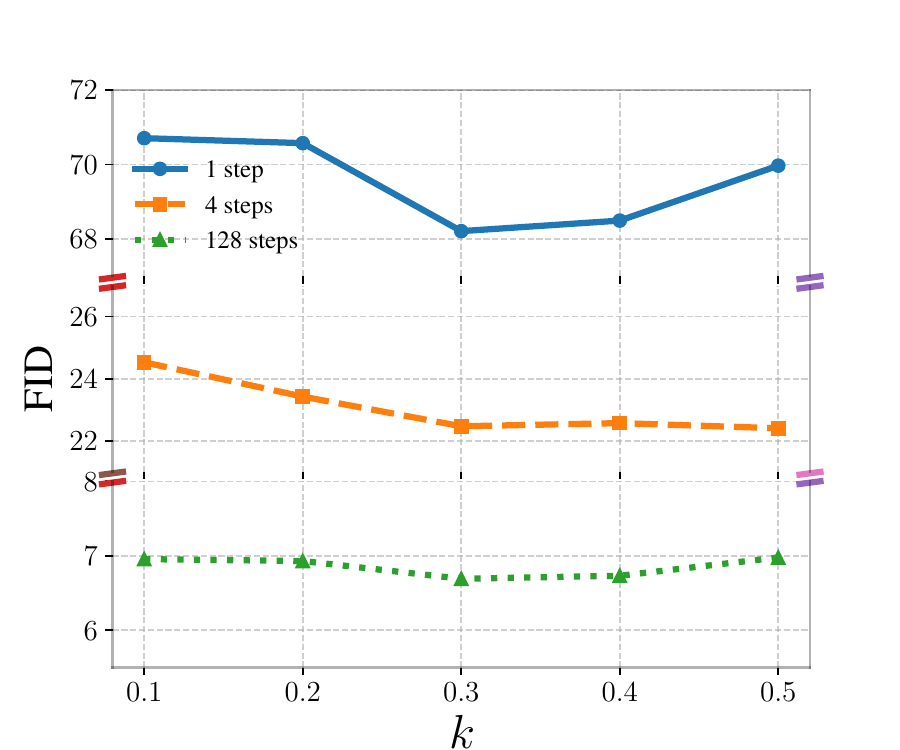}}
\subfloat[Hyper-parameter $r$]{
\includegraphics[width=0.32\textwidth]{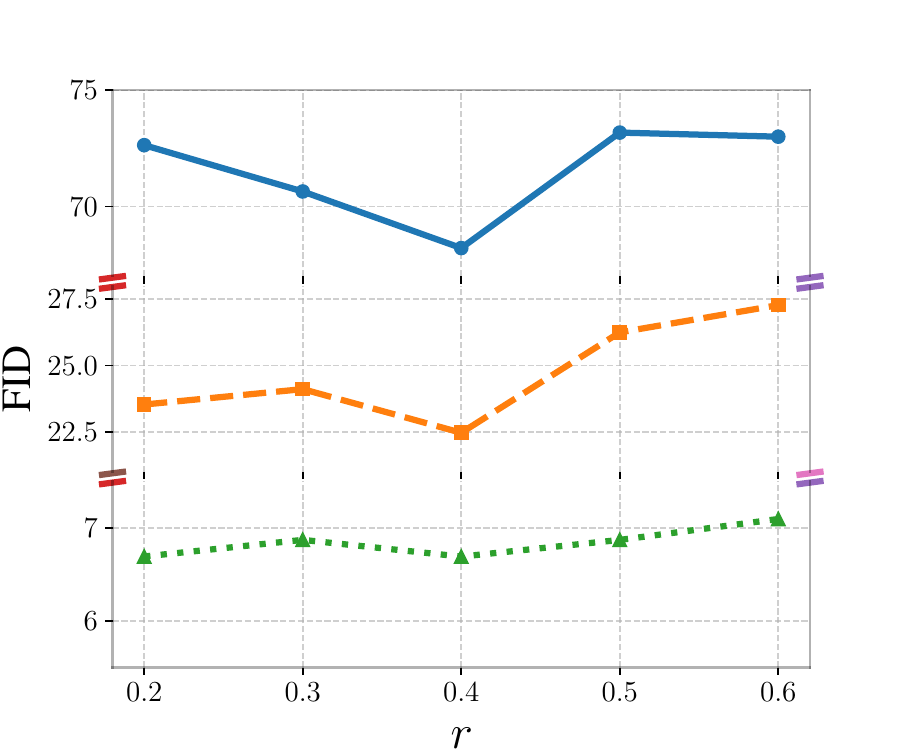}}
\subfloat[Hyper-parameter $\lambda$]{
\includegraphics[width=0.32\textwidth]{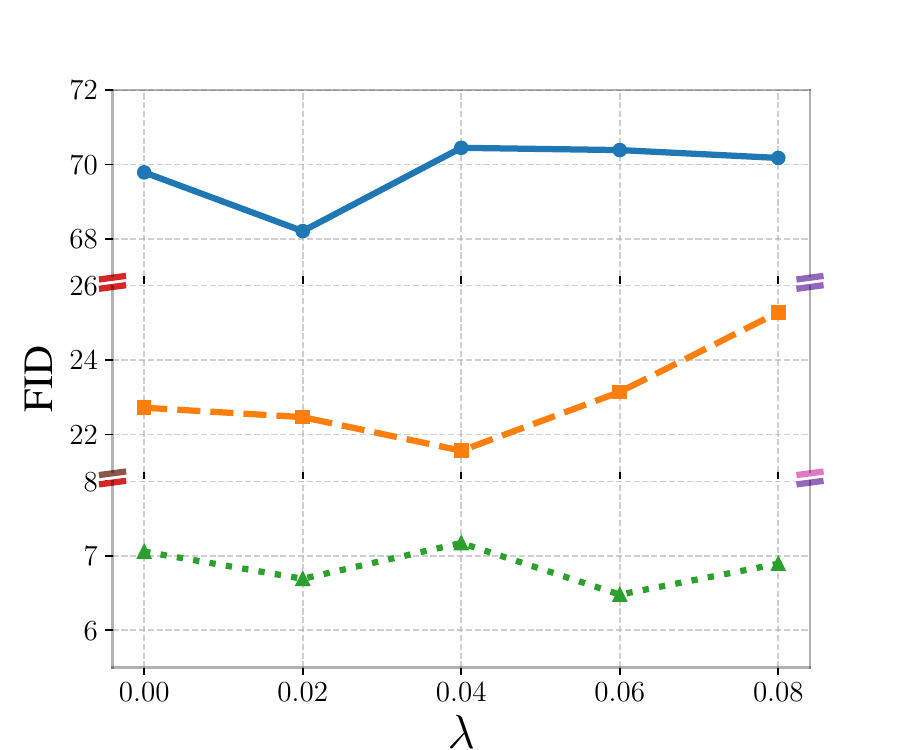}}
\caption{Sensitivity analysis on MNIST for hyperparameters $k$, $r$, and $\lambda$.}
\label{fig:sen}
\vspace{-10px}
\end{figure}

As shown in Tab.~\ref{tab:fid_results}, MAC consistently improves sample quality across datasets, especially under few-step settings. On MNIST, our method achieves a 6.8 FID reduction over Shortcut in the one-step setting. On CIFAR-10, MAC improves performance over Shortcut in both the one-step and 128-step settings. CelebA-HQ results show similar trends, where MAC achieves the lowest FID in the one-step and four-step settings. These improvements validate that MAC enables the model to focus on learnable transport directions, resulting in more efficient generation.

\subsection{Hyperparameter Sensitivity}

To assess the sensitivity of MAC and guide hyperparameter selection, we conduct a sensitivity analysis on three key hyperparameters: the selection ratio $k$, the one-step supervision ratio $r$, and the weighting coefficient $\lambda$. Specifically, $k$ controls the proportion of low-prediction-error couplings used for regularization, $r$ determines the fraction of selected couplings used to train the model with a fixed step size $d = 1$, and $\lambda$ balances the regularization strength. We evaluate $k \in \{0.1,\dots,0.5\}$, $r \in \{0.2,\dots,0.6\}$, and $\lambda \in \{0,\dots,0.08\}$, and report 1-step, 4-step, and 128-step FID scores on MNIST As shown in Fig.~\ref{fig:sen}), MAC is not sensitive to the hyperparameters. Based on these results, we select $k = 0.3$, $r = 0.4$, and $\lambda = 0.02$ as default settings in our experiments.

\begin{figure}[t]
\centering
\subfloat[Shortcut]{
\includegraphics[width=0.43\textwidth]{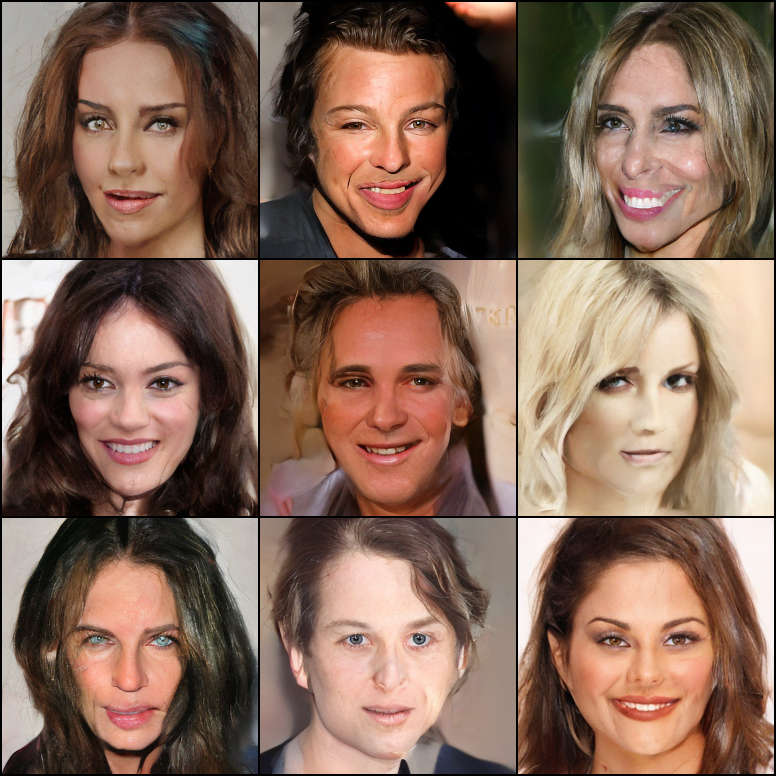}} \quad
\subfloat[MAC (Ours)]{
\includegraphics[width=0.43\textwidth]{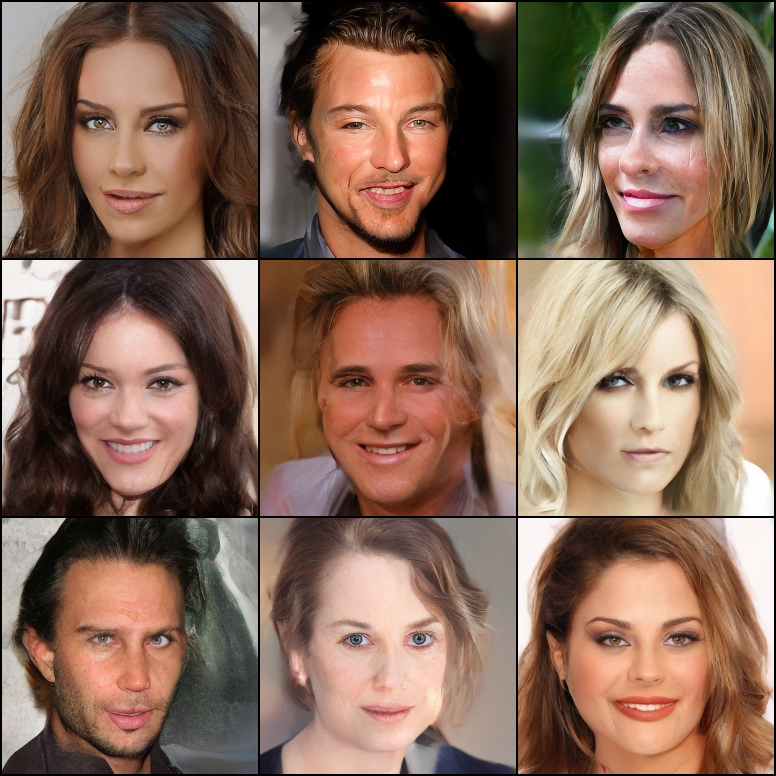}}
\caption{Qualitative comparison on CelebA-HQ-256 under the four-step generation setting. Compared to Shortcut Models, MAC generates images with more realistic facial details, smoother textures, and better global structure.}
\label{fig:qualitative}
\vspace{-10px}
\end{figure}

\subsection{Qualitative Results}

Fig.~\ref{fig:qualitative} presents qualitative results on CelebA-HQ-256 under the four-step generation setting. While both methods aim to synthesize realistic images with limited sampling steps, MAC consistently produces samples with higher perceptual fidelity. In particular, MAC yields sharper facial details, more coherent structural features (e.g., mouth, teeth, and facial contours), and significantly fewer visual artifacts compared to Shortcut Models. These results highlight MAC's ability to guide the model toward straighter transport trajectories, ultimately leading to improved generation quality in few-step settings.

\section{Conclusion}

We presented Model-Aligned Coupling (MAC), a simple yet effective framework for improving flow-based generative models by selecting learnable couplings during training. Unlike existing methods that construct couplings by solely considering geometry distance, MAC leverages the model's evolving vector field to identify couplings with low prediction error, thereby aligning supervision with directions the model can more effectively learn. We demonstrate that MAC can be seamlessly integrated as a regularization method into a wide range of flow matching objectives, including the Shortcut Models. Extensive experiments across multiple datasets show that MAC significantly improves generation quality, particularly under few-step settings such as one-step and four-step generation.
These results underscore the importance of coupling strategies that adapt to the model’s vector field, and suggest that constructing training couplings based on the model’s capacity is a promising direction for future research in flow-based generative modeling.

{
\small
\bibliographystyle{plain}
\bibliography{main.bib}

\begin{thebibliography}{10}

\bibitem{cuturi2013sinkhorn}
Marco Cuturi.
\newblock Sinkhorn distances: Lightspeed computation of optimal transport.
\newblock {\em Advances in neural information processing systems}, 26, 2013.

\bibitem{feng2024relational}
Weilun Feng, Chuanguang Yang, Zhulin An, Libo Huang, Boyu Diao, Fei Wang, and Yongjun Xu.
\newblock Relational diffusion distillation for efficient image generation.
\newblock In {\em Proceedings of the 32nd ACM International Conference on Multimedia}, pages 205--213, 2024.

\bibitem{frans2024one}
Kevin Frans, Danijar Hafner, Sergey Levine, and Pieter Abbeel.
\newblock One step diffusion via shortcut models.
\newblock {\em arXiv preprint arXiv:2410.12557}, 2024.

\bibitem{ho2020denoising}
Jonathan Ho, Ajay Jain, and Pieter Abbeel.
\newblock Denoising diffusion probabilistic models.
\newblock {\em Advances in neural information processing systems}, 33:6840--6851, 2020.

\bibitem{karras2017progressive}
Tero Karras, Timo Aila, Samuli Laine, and Jaakko Lehtinen.
\newblock Progressive growing of gans for improved quality, stability, and variation.
\newblock {\em arXiv preprint arXiv:1710.10196}, 2017.

\bibitem{karras2022elucidating}
Tero Karras, Miika Aittala, Timo Aila, and Samuli Laine.
\newblock Elucidating the design space of diffusion-based generative models.
\newblock {\em Advances in neural information processing systems}, 35:26565--26577, 2022.

\bibitem{krizhevsky2009learning}
Alex Krizhevsky, Geoffrey Hinton, et~al.
\newblock Learning multiple layers of features from tiny images.
\newblock 2009.

\bibitem{lecun1998gradient}
Yann LeCun, L{\'e}on Bottou, Yoshua Bengio, and Patrick Haffner.
\newblock Gradient-based learning applied to document recognition.
\newblock {\em Proceedings of the IEEE}, 86(11):2278--2324, 1998.

\bibitem{lee2024improving}
Sangyun Lee, Zinan Lin, and Giulia Fanti.
\newblock Improving the training of rectified flows.
\newblock {\em Advances in Neural Information Processing Systems}, 37:63082--63109, 2024.

\bibitem{lipman2022flow}
Yaron Lipman, Ricky~TQ Chen, Heli Ben-Hamu, Maximilian Nickel, and Matt Le.
\newblock Flow matching for generative modeling.
\newblock {\em arXiv preprint arXiv:2210.02747}, 2022.

\bibitem{liu2022flow}
Xingchao Liu, Chengyue Gong, and Qiang Liu.
\newblock Flow straight and fast: Learning to generate and transfer data with rectified flow.
\newblock {\em arXiv preprint arXiv:2209.03003}, 2022.

\bibitem{lu2022dpm}
Cheng Lu, Yuhao Zhou, Fan Bao, Jianfei Chen, Chongxuan Li, and Jun Zhu.
\newblock Dpm-solver: A fast ode solver for diffusion probabilistic model sampling in around 10 steps.
\newblock {\em Advances in Neural Information Processing Systems}, 35:5775--5787, 2022.

\bibitem{peebles2023scalable}
William Peebles and Saining Xie.
\newblock Scalable diffusion models with transformers.
\newblock In {\em Proceedings of the IEEE/CVF international conference on computer vision}, pages 4195--4205, 2023.

\bibitem{pooladian2023multisample}
Aram-Alexandre Pooladian, Heli Ben-Hamu, Carles Domingo-Enrich, Brandon Amos, Yaron Lipman, and Ricky~TQ Chen.
\newblock Multisample flow matching: Straightening flows with minibatch couplings.
\newblock {\em arXiv preprint arXiv:2304.14772}, 2023.

\bibitem{qu2025spatialvlaexploringspatialrepresentations}
Delin Qu, Haoming Song, Qizhi Chen, Yuanqi Yao, Xinyi Ye, Yan Ding, Zhigang Wang, JiaYuan Gu, Bin Zhao, Dong Wang, and Xuelong Li.
\newblock Spatialvla: Exploring spatial representations for visual-language-action model, 2025.

\bibitem{salimans2022progressive}
Tim Salimans and Jonathan Ho.
\newblock Progressive distillation for fast sampling of diffusion models.
\newblock {\em arXiv preprint arXiv:2202.00512}, 2022.

\bibitem{sohl2015deep}
Jascha Sohl-Dickstein, Eric Weiss, Niru Maheswaranathan, and Surya Ganguli.
\newblock Deep unsupervised learning using nonequilibrium thermodynamics.
\newblock In {\em International conference on machine learning}, pages 2256--2265. pmlr, 2015.

\bibitem{song2025humeintroducingsystem2thinking}
Haoming Song, Delin Qu, Yuanqi Yao, Qizhi Chen, Qi~Lv, Yiwen Tang, Modi Shi, Guanghui Ren, Maoqing Yao, Bin Zhao, Dong Wang, and Xuelong Li.
\newblock Hume: Introducing system-2 thinking in visual-language-action model, 2025.

\bibitem{song2020denoising}
Jiaming Song, Chenlin Meng, and Stefano Ermon.
\newblock Denoising diffusion implicit models.
\newblock {\em arXiv preprint arXiv:2010.02502}, 2020.

\bibitem{song2023consistency}
Yang Song, Prafulla Dhariwal, Mark Chen, and Ilya Sutskever.
\newblock Consistency models.
\newblock 2023.

\bibitem{song2020score}
Yang Song, Jascha Sohl-Dickstein, Diederik~P Kingma, Abhishek Kumar, Stefano Ermon, and Ben Poole.
\newblock Score-based generative modeling through stochastic differential equations.
\newblock {\em arXiv preprint arXiv:2011.13456}, 2020.

\bibitem{song2023equivariant}
Yuxuan Song, Jingjing Gong, Minkai Xu, Ziyao Cao, Yanyan Lan, Stefano Ermon, Hao Zhou, and Wei-Ying Ma.
\newblock Equivariant flow matching with hybrid probability transport for 3d molecule generation.
\newblock {\em Advances in Neural Information Processing Systems}, 36:549--568, 2023.

\bibitem{su2024roformer}
Jianlin Su, Murtadha Ahmed, Yu~Lu, Shengfeng Pan, Wen Bo, and Yunfeng Liu.
\newblock Roformer: Enhanced transformer with rotary position embedding.
\newblock {\em Neurocomputing}, 568:127063, 2024.

\bibitem{tong2023improving}
Alexander Tong, Kilian Fatras, Nikolay Malkin, Guillaume Huguet, Yanlei Zhang, Jarrid Rector-Brooks, Guy Wolf, and Yoshua Bengio.
\newblock Improving and generalizing flow-based generative models with minibatch optimal transport.
\newblock {\em arXiv preprint arXiv:2302.00482}, 2023.

\bibitem{tong2023simulation}
Alexander Tong, Nikolay Malkin, Kilian Fatras, Lazar Atanackovic, Yanlei Zhang, Guillaume Huguet, Guy Wolf, and Yoshua Bengio.
\newblock Simulation-free schr$\backslash$" odinger bridges via score and flow matching.
\newblock {\em arXiv preprint arXiv:2307.03672}, 2023.

\bibitem{wang2024rectified}
Fu-Yun Wang, Ling Yang, Zhaoyang Huang, Mengdi Wang, and Hongsheng Li.
\newblock Rectified diffusion: Straightness is not your need in rectified flow.
\newblock {\em arXiv preprint arXiv:2410.07303}, 2024.

\bibitem{wildberger2023flow}
Jonas Wildberger, Maximilian Dax, Simon Buchholz, Stephen Green, Jakob~H Macke, and Bernhard Sch{\"o}lkopf.
\newblock Flow matching for scalable simulation-based inference.
\newblock {\em Advances in Neural Information Processing Systems}, 36:16837--16864, 2023.

\bibitem{zhang2025towards}
Yichi Zhang, Yici Yan, Alex Schwing, and Zhizhen Zhao.
\newblock Towards hierarchical rectified flow.
\newblock {\em arXiv preprint arXiv:2502.17436}, 2025.

\end{thebibliography}
}

% \section*{References}

% References follow the acknowledgments in the camera-ready paper. Use unnumbered first-level heading for
% the references. Any choice of citation style is acceptable as long as you are
% consistent. It is permissible to reduce the font size to \verb+small+ (9 point)
% when listing the references.
% Note that the Reference section does not count towards the page limit.
% \medskip

% {
% \small

% [1] Alexander, J.A.\ \& Mozer, M.C.\ (1995) Template-based algorithms for
% connectionist rule extraction. In G.\ Tesauro, D.S.\ Touretzky and T.K.\ Leen
% (eds.), {\it Advances in Neural Information Processing Systems 7},
% pp.\ 609--616. Cambridge, MA: MIT Press.

% [2] Bower, J.M.\ \& Beeman, D.\ (1995) {\it The Book of GENESIS: Exploring
%   Realistic Neural Models with the GEneral NEural SImulation System.}  New York:
% TELOS/Springer--Verlag.

% [3] Hasselmo, M.E., Schnell, E.\ \& Barkai, E.\ (1995) Dynamics of learning and
% recall at excitatory recurrent synapses and cholinergic modulation in rat
% hippocampal region CA3. {\it Journal of Neuroscience} {\bf 15}(7):5249-5262.
% }

%%%%%%%%%%%%%%%%%%%%%%%%%%%%%%%%%%%%%%%%%%%%%%%%%%%%%%%%%%%%

\newpage

\appendix

\section{Extra Experiment Results}
\label{app:extra}

\begin{table}[h]
\centering
\caption{
FID comparison across one-step, four-step, and 128-step generation on MNIST, CIFAR-10, and CelebA-HQ-256. 
Lower is better. Best results in each group are highlighted in bold.
}

\label{tab:fid_results_appendix}
\resizebox{\linewidth}{!}{
\begin{tabular}{l|ccc|ccc|ccc}
\toprule
\multirow{2}{*}{Method} & \multicolumn{3}{c|}{MNIST (Cond.)} & \multicolumn{3}{c|}{CIFAR-10 (Cond.)} & \multicolumn{3}{c}{CelebA-HQ (Uncond.)} \\
& 1-Step & 4-Step & 128-Step & 1-Step & 4-Step & 128-Step & 1-Step & 4-Step & 128-Step \\
\midrule
Shortcut      &  75.03 & 25.80 & 7.22 &  36.19 & 17.14 & 11.96 & 32.20 & 17.76 & 11.16 \\
MAC (ours)    &  68.21 & 22.47 & 6.69 & 35.47 & 19.14 & 10.44 & 26.48 & 13.84 & 11.40 \\
MAC-full (ours)    &  \textbf{67.82} & \textbf{18.23} & \textbf{5.50} & \textbf{33.75} & \textbf{16.18} & \textbf{10.42} & \textbf{22.76} & \textbf{13.43} & \textbf{9.86} \\
\bottomrule
\end{tabular}
}
\end{table}

\paragraph{Full Coupling Optimization.}
In the main paper, we adopt a top-$k$ selection strategy to approximate the ideal coupling $\tilde{\rho}$ defined in Eq.~\ref{eq:rho}, by selecting a fixed fraction of source-target couplings with the lowest prediction error. Here, we investigate a more accurate alternative that performs \textit{full coupling optimization} by solving Eq.~\ref{eq:rho} over all possible source-target couplings within each batch. Specifically, we solve the full assignment problem defined in Eq.~\ref{eq:rho} by applying the Sinkhorn algorithm \cite{cuturi2013sinkhorn} with the regularization parameter $\texttt{reg}=0.5$ to minimize the expected pairwise prediction error over all source-target couplings. The resulting couplings is then used to train a shortcut model \cite{frans2024one}. We refer to this variant as MAC-full.

While full coupling introduces additional computation, we find that its training cost remains comparable to that of the top-$k$ variant. For instance, on CIFAR-10, MAC (Top-$k$ Selection) completes training in 5h06m, whereas MAC-full requires 5h13m. As shown in Tab.~\ref{tab:fid_results_appendix}, the full coupling variant consistently achieves better performance across datasets and sampling steps, highlighting the benefits of more precise pairwise supervision.

\begin{table}[h]
\centering
\caption{FID comparison across one-step, four-step, and 128-step generation on MNIST, CIFAR-10, and CelebA-HQ-256. Lower is better. The best results within each category are highlighted in bold.}
\label{tab:fid_results_extra}
\resizebox{\linewidth}{!}{
\begin{tabular}{l|ccc|ccc|ccc}
\toprule
\multirow{2}{*}{Method} & \multicolumn{3}{c|}{MNIST (Cond.)} & \multicolumn{3}{c|}{CIFAR-10 (Cond.)} & \multicolumn{3}{c}{CelebA-HQ (Uncond.)} \\
& 1-Step & 4-Step & 128-Step & 1-Step & 4-Step & 128-Step & 1-Step & 4-Step & 128-Step \\
\midrule
Flow Matching & 169.97 & \textbf{28.99}& 7.13 & 324.04 & 36.85 & 11.05 & \textbf{284.99} & 65.05 & 15.25 \\
FM + MAC-full (ours)    &  \textbf{165.58} & 29.90 & \textbf{6.71} & \textbf{311.52} & \textbf{35.61} & \textbf{10.39} & 285.19 & \textbf{64.69} & \textbf{13.39} \\
\bottomrule
\end{tabular}
}
\end{table}

\paragraph{Improving Flow Matching via Model-Aligned Coupling.}
To demonstrate the general applicability of our method, we integrate MAC-full with the standard Flow Matching (FM) objective, replacing its original random coupling strategy with our full coupling optimization. We denote this variant as FM + MAC-full. As shown in Tab.~\ref{tab:fid_results_extra}, incorporating MAC-full consistently improves FID scores across all datasets and sampling steps compared to vanilla Flow Matching. This result highlights our method can enhance a wide range of existing flow-based generative models by providing more learnable and aligned supervision.

\section{More Implementation Details}
\label{app:details}
\paragraph{Network Structure}

\begin{table}[h]
\centering
\caption{DiT model configurations used for different datasets. All models share a common Transformer-based architecture with AdaLN conditioning. Minor differences lie in positional encoding schemes and input embedding strategies.}
\begin{tabular}{lccc}
\toprule
\textbf{Configuration} & \textbf{MNIST} & \textbf{CIFAR-10} & \textbf{CelebA-HQ-256} \\
\midrule
Patch Size             & 2                 & 2                 & 2 \\
Depth                  & 6                 & 10                & 12 \\
Hidden Size            & 64                & 256               & 768 \\
Number of Heads        & 4                 & 8                 & 12 \\
Input Shape            & (1, 32, 32)       & (3, 32, 32)       & (4, 32, 32) \\
Class Conditioning     & Yes          & Yes               & No \\
Positional Encoding    & Rotary Embedding  & Rotary Embedding  & 2D Sin-Cos Embedding \\
Input Embedding               & CNN + GroupNorm   & CNN + GroupNorm   & PatchEmbed \\
\bottomrule
\end{tabular}

\label{tab:dit_configs}
\end{table}

We adopt the DiT architecture~\cite{peebles2023scalable} as a unified backbone across all experiments. Tab.~\ref{tab:dit_configs} summarizes the specific model configurations for each dataset. For MNIST and CIFAR-10, we use a DiT model with a convolutional input embedding, consisting of two convolutional layers with SiLU activations and Group Normalization, followed by a linear projection of image patches. Rotary positional embeddings \cite{su2024roformer} are applied to encode spatial information. This configuration is particularly effective for low-resolution images (e.g., $32 \times 32$), where convolutional layers can better capture local patterns.

For CelebA-HQ-256, we use a standard DiT \cite{peebles2023scalable} model with patch-based input embedding and fixed 2D sinusoidal positional encodings, as commonly used in vision Transformers. While the input embedding modules and positional encoding strategies differ across datasets, all models share the same Transformer backbone with AdaLN conditioning and are trained using the same generative objective.

\begin{figure}[h]
\vspace{-10px}
\centering
\includegraphics[width=0.9\textwidth,trim=0 170 310 0,clip]{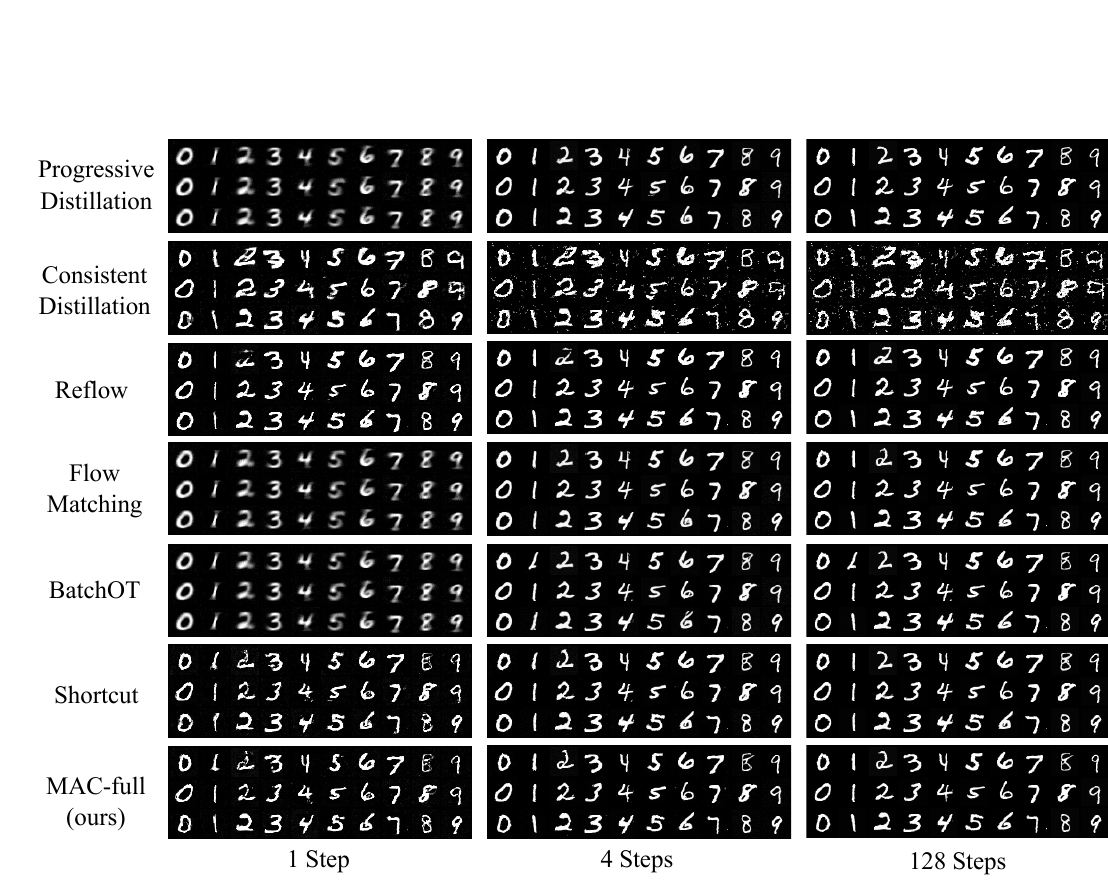}
\vspace{-10px}
\caption{Qualitative comparison of generated MNIST samples across different models under varying sampling steps (1, 4, and 128 steps). MAC-full generates sharper digits, especially in the one-step and four-step settings.}
\label{fig:vis_mnist}
\vspace{-0px}
\end{figure}

\begin{figure}[h]
\vspace{-10px}
\centering
\includegraphics[width=0.9\textwidth,trim=10 170 300 0,clip]{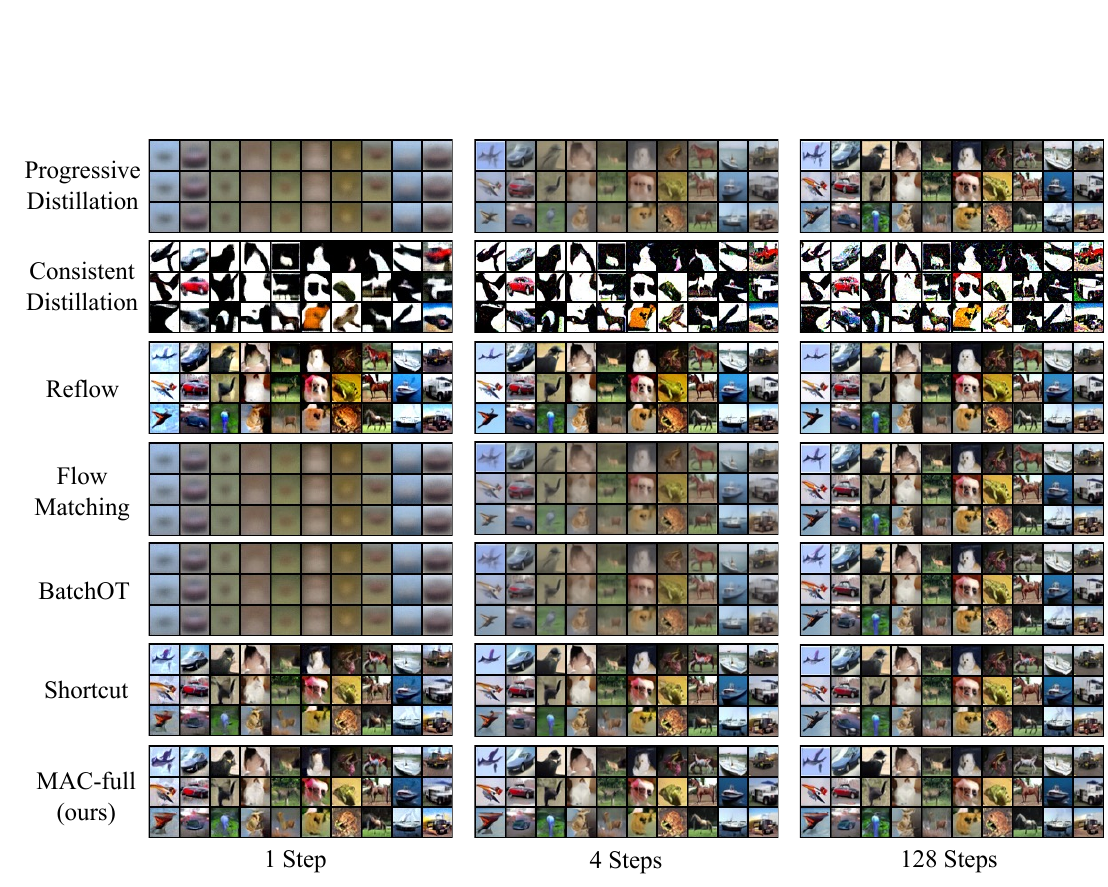}
\vspace{-7px}
\caption{Qualitative comparison of generated CIFAR-10 samples across different models under varying sampling steps (1, 4, and 128 steps). MAC-full generates more realistic and structurally consistent objects, especially in the one-step and four-step settings.}
\label{fig:vis_cifar}
\vspace{-10px}
\end{figure}

\begin{figure}[h]
\vspace{-10px}
\centering
\includegraphics[width=1\textwidth,trim=0 95 410 0,clip]{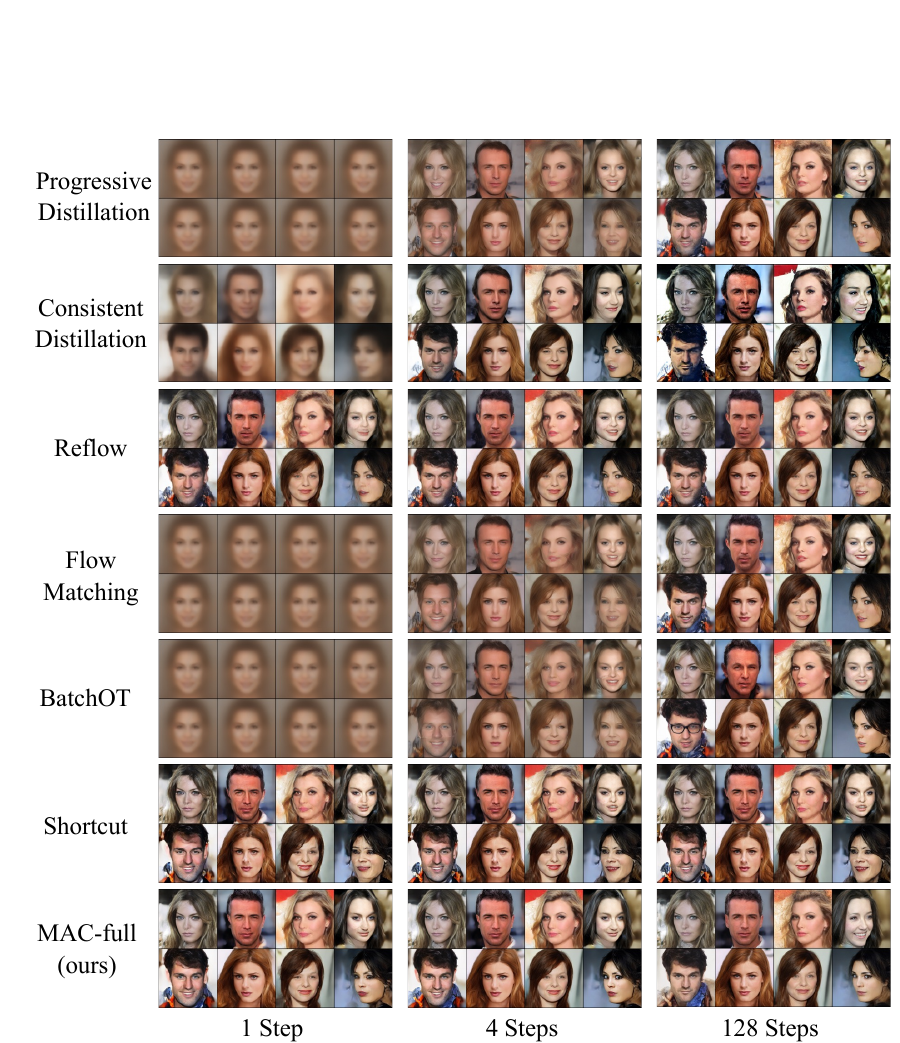}
\vspace{-10px}
\caption{Qualitative comparison of generated CelebA-HQ-256 samples across different models under varying sampling steps (1, 4, and 128 steps). MAC-full shows superior facial details and fewer artifacts, especially in the one-step and four-step settings.}
\label{fig:vis_celeba}
\vspace{-10px}
\end{figure}

\section{More Qualitative Results}

To further validate the effectiveness of our proposed method, we provide qualitative comparisons of generated samples across different baselines and sampling steps. Specifically, we visualize samples from the following models: Progressive Distillation, Consistent Distillation, Reflow, Flow Matching, BatchOT, Shortcut, and our method MAC-full. For each model, we show generated results under three sampling regimes: one-step, four-step, and full (128-step) generation.

Fig.~\ref{fig:vis_mnist}, \ref{fig:vis_cifar}, and \ref{fig:vis_celeba} present qualitative results on MNIST, CIFAR-10, and CelebA-HQ-256, respectively. Across all datasets and sampling steps, MAC-full consistently produces samples with the highest perceptual quality, exhibiting sharper details, more coherent structures, and fewer visual artifacts compared to prior methods. Notably, the improvement is especially pronounced in the low-step settings (e.g., 1-step and 4-step), where other methods tend to produce blurry or distorted outputs, while MAC-full maintains high visual fidelity.

These results qualitatively support our findings and demonstrate that model-aligned coupling significantly improves sample quality, particularly in few-step generation scenarios.

\section{Limitations}

A limitation is the additional computational overhead caused by the need to evaluate the model's prediction error during training. Specifically, both the top-$k$ selection and full coupling optimization variants require extra forward passes through the model to compute velocity predictions at $t = 0$ and $t = 1$, which increases training cost compared to standard Flow Matching. 

\section{Social Impact and Safeguards}
This work proposes a technique to improve the efficiency and quality of sample generation in flow-based generative models. As such, it does not introduce new forms of data collection, sensitive attribute usage, or deployment risks beyond those already present in existing generative modeling frameworks. However, like many generative methods, MAC could be applied to produce synthetic images that may be misused in misinformation or deepfake contexts. We stress that our contributions are algorithmic in nature and not tied to any specific downstream application. To mitigate potential misuse, we recommend that models trained using MAC be accompanied by usage guidelines, watermarks, or authenticity verification tools when deployed in open-ended generation settings.

\end{document}